# Optimizing Causal Orderings for Generating DAGs from Data

**Remco R. Bouckaert**
Utrecht University
Department of Computer Science
P.O.Box 80.089 3508 TB Utrecht,The Netherlands
remco@cs.ruu.nl

## Abstract

An algorithm for generating the structure of a directed acyclic graph from data using the notion of causal input lists is presented. The algorithm manipulates the ordering of the variables with operations which very much resemble arc reversal. Operations are only applied if the DAG after the operation represents at least the independencies represented by the DAG before the operation until no more arcs can be removed from the DAG. The resulting DAG is a minimal I-map.

## 1 Introduction

Bayesian belief networks (BBN) can very well be used as models for real world problems which deal with uncertainty. Most methods for modeling a domain with a BBN [3, 5, 7] however, are not capable of capturing all the dependencies which are in the domain. Therefore, the resulting BBN will contain unnecessary errors and the algorithms for calculating beliefs [4, 7] will not provide exact answers. On the other hand, it is easy to construct a BBN that does represent all dependencies of the domain. A fully connected graph is an extreme example of such a BBN.

With BBNs most of the time chordal graphs or directed acyclic graphs (DAG) are meant. In this paper we present an algorithm that generates the structure of a DAG which represents all dependencies in the data. The algorithm uses the notion of causal input lists. A causal input list fixes a DAG by an ordering on the variables in the domain. Operations are defined on this ordering such that the set of independencies represented by the DAG is monotonically increasing.

So, by applying these operations the corresponding model will converge to an optimal DAG. This DAG is optimal in the sense that no arcs can be deleted by applying the operations on the ordering.

In section 2 we explains terms and definitions and the strategy we follow for finding an optimal ordering. In the sections 3, 4, 5 and 6 we describe operations on orderings. These operations are used for the algorithm which finds an optimal ordering on the variables. This algorithm is described in section 7.

## 2 Preliminaries

The goal is to represent a probability distribution over a set of variables, we call $U$, with a DAG. With every variable $u \in U$ a node in the DAG is associated. In this paper we write capital letters to denote sets of variables and lower case letters to denote single variables. For the set of nodes (or single node) corresponding to a set of variables (or single variable) we will use the same name. All variables or sets of variables mentioned are elements or subsets of $U$ unless stated otherwise.

The structure of the DAG should represent the independencies in the distribution. We call $X$, $Y$ *conditionally independent* given $Z$, written $I(X, Z, Y)$, if $X$ is statistically independent from $Y$ given $Z$. $I(X, Z, Y)$ is an *independency statement*. An *independency model* $M$ over $U$ is a set of independency statements $I(X, Z, Y)$ with $X, Y, Z \subseteq U$. A *complete independency model* $M$ of a distribution over $U$ is the set of all valid independency statements in this distribution. In this paper we assume that the distribution is positive definite ($P > 0$) and it can be verified for any statement $I(X, Z, Y)$ if it is in $M$. Therefore the rules of inference called symmetry ($I(X, Z, Y) \Leftrightarrow I(Y, Z, X)$), decomposition ($I(X, Z, WY) \Rightarrow I(X, Z, W)$), weak union ($I(X, Z, WY) \Rightarrow I(X, ZW, Y)$), contraction ($I(X, Z, Y) \wedge I(X, ZY, W) \Rightarrow I(X, Z, WY)$) and intersection ($I(X, ZW, Y) \wedge I(X, ZY, W) \Rightarrow I(X, Z, YW)$) apply.

Independency statements in the distribution can be

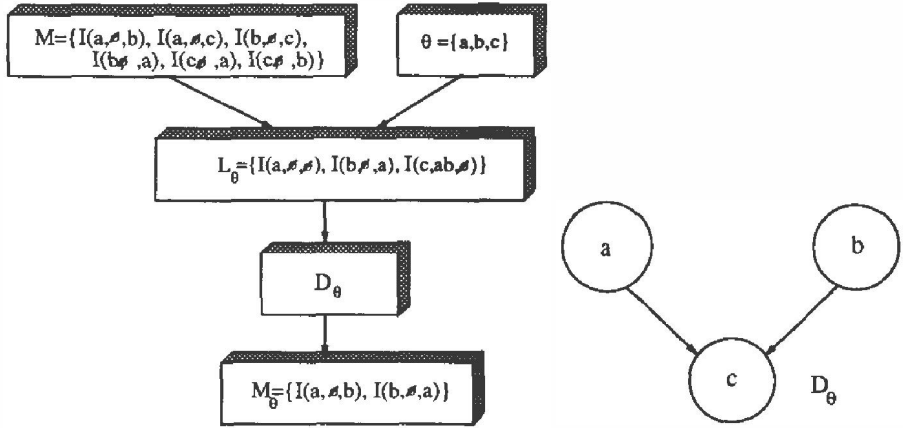

Figure 1: Example showing the relation between $M$, $\theta$, $L_\theta$, $D_\theta$ and $M_\theta$

read from the structure of the DAG with the notion of separation. A *head-to-head node* in a DAG $D$ is a triple nodes $a, b, c$ such that $a \rightarrow b \leftarrow c$ not $a \rightarrow c$ nor $c \rightarrow a$ in $D$. $X$ is *separated* from $Y$ given $Z$, written $<X, Z, Y>$, if for every undirected path from every node in $X$ to every node in $Y$ at least one of the next two cases hold:

1: The path contains a head-to-head node $a \rightarrow b \leftarrow c$ and $b \notin Z$ and every descendant of $b$ is not in $Z$.

2: There is a node $b$ in the path with $b \in Z$ and $b$ is not in a head-to-head node $a \rightarrow b \leftarrow c$ in the path.

A DAG can be constructed from a complete independency model $M$ using the notion of causal input lists [8].

**Definition 2.1** *Let $\theta$ be a total ordering over the set of variables $U$ then $\theta$ is a* causal ordering. *A causal input list $L_\theta$ over a complete independency model $M$ is set of independency statements such that for every $u \in U$, $L_\theta$ contains exactly one independency statement which has format:*

$$T = I(u, B_u, U_u \backslash B_u)$$

*in which $U_u = \{v|v \in U, \theta(v) < \theta(u)\}$ and $B_u$ is the smallest subset of $U_u$ such that $T$ holds. $B_u$ is called the* boundary *of $u$.*

A DAG $D_\theta$ can be associated with the ordering $\theta$ by placing an arc for each node $u$ from every node in its boundary to node $u$. Likewise an independency model $M_\theta$ can be associated with this ordering by letting $I(X, Z, Y) \in M_\theta$ iff $<X, Z, Y>$ holds in $D_\theta$. Now it is known $M_\theta \subseteq M$ for any $\theta$ [8].

To illustrate the just mentioned notions, consider the situation the Netherlands and Germany are playing the semi final soccer against opponents which are two other national teams of equal strength. Let $U = \{a, b, c\}$ and $a$ is the variable representing the Netherlands will win, $b$ that Germany will win and $c$ that the Netherlands will meet in the final for the first and second place or for the third and fourth place. So, if both the Netherlands and Germany win or both loose they will meet. Since it is possible one of the parties has to quit due to poisoning by food, the distribution is positive definite. Figure 1 shows the complete independency model $M$ we assume to hold for these variables. From the ordering $\theta = \{a, b, c\}$ and $M$ the causal input list $L_\theta$ is generated. The associated DAG $D_\theta$ is shown and its independency model $M_\theta$. Remark $I(a, \emptyset, c)$ is not in $M_\theta$ though it is in $M$!

A DAG $D$ is an *I-map* of a complete independency model $M$ if $<X, Z, Y> \Rightarrow I(X, Z, Y) \in M$ and it is a *minimal I-map* if no arc can be removed without destroying its I-mappedness. It is a *D-map* if $I(X, Z, Y) \in M \Rightarrow <X, Z, Y>$ and it is a *perfect map* if it is both an I-map and a D-map. A complete independency model $M$ is *DAG-isomorph* if there exists a DAG which is a perfect map of $M$.

Let $\Theta$ be the set of all possible orderings. To find an ordering $\theta$ such that $M_\theta = M$ we step through the solution space $\Theta$. A step from an ordering $\theta$ to $\theta'$ is only made if $M_\theta \subseteq M_{\theta'}$. This way the represented model monotonically increases in size and we get a path of orderings $\theta_0, \theta_1, ..., \theta_n$ with corresponding models $M_{\theta_0}, M_{\theta_1}, ..., M_{\theta_n}$ which converges to $M$ if $M$ is DAG-isomorph (see Fig. 2).

An ordering $\theta$ is *reducible* if an ordering $\theta'$ exists with $M_\theta \subset M_{\theta'}$. A clique in $D_\theta$ is called reducible if an ordering $\theta'$ exists such that at least one arc in the clique can be removed.

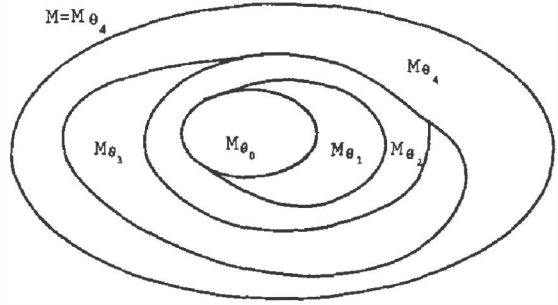

Figure 2: Effect of search-strategy.

## 3 The swap operator

A step in $\Theta$ can be done with the swap operator. The swap operator, written $\theta' = swap(\theta, \theta(i), \theta(j))$, changes the order of two consecutive nodes $i$ and $j$ (see Fig. 3). Using the swap operator one can *shift* a node to any place in the ordering one wants (see Fig. 4). So any $\theta \in \Theta$ can be reached using the swap operator. In this section we will see what the consequences are for the boundaries in the causal input list. This very much resembles the so-called arc reversal as described in [6, 9].

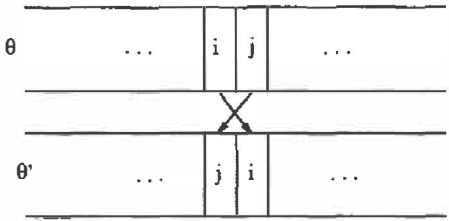

Figure 3: Effect of swap on an ordering $\theta$.

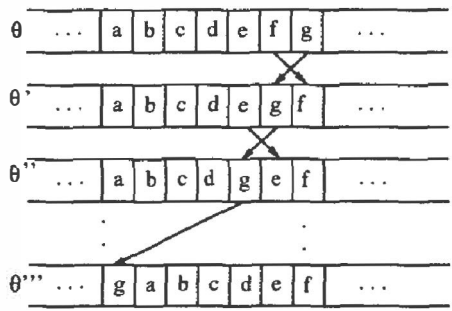

Figure 4: Shifting a node through the ordering $\theta$ by swapping.

**Lemma 3.1** *Let $M$ be a complete independency model over $U$. Let $L_\theta$ be a causal input list of $M$ with $B_a$ the boundaries in $L_\theta$. Let $i, j \in U$ be two variables with* $\theta(i) = \theta(j) - 1$. *Let $\theta' = swap(\theta, \theta(i), \theta(j))$ with causal input list $L_{\theta'}$ and $B'_a$ the boundaries in $L_{\theta'}$ then*

$$B'_i \cup B'_j \setminus \{j\} = B_i \cup B_j \setminus \{i\}$$

Sketch of the proof: For $i \notin B_j$ it is trivial so we regard the case $i \in B_j$. First we show $B'_i \cup B'_j \setminus \{j\} \subseteq B_i \cup B_j \setminus \{i\}$. From $L_\theta$ we have

$$I(i, B_i, U_i \setminus B_i) \wedge I(j, B_j, U_j \setminus B_j)$$
$$= \{ A = B_i \setminus B_j, B = B_i \cap B_j \text{ and } C = B_j \setminus (B_i \cup \{i\}) \}$$
$$I(i, AB, U_i \setminus AB) \wedge I(j, BC\{i\}, U_i \setminus BC)$$
$$\Rightarrow \{ \text{Weak Union} \}$$
$$I(i, ABC, U_i \setminus ABC) \wedge I(j, ABC\{i\}, U_i \setminus ABC)$$
$$\Rightarrow \{ \text{Contraction} \}$$
$$I(ij, ABC, U_i \setminus ABC)$$
$$\Rightarrow \{ \text{Weak Union and Decomposition} \}$$
$$I(i, ABC\{j\}, U_i \setminus ABC) \wedge I(j, ABC, U_i \setminus ABC)$$

By assuming the boundary of $i$ has nodes outside $ABC\{j\}$ using intersection one can derive a contradiction. Likewise for $B'_j$. Therefore $B'_i \subseteq ABC\{j\}$ and $B'_j \subseteq ABC$ so $B'_i \cup B'_j \setminus \{i\} \subseteq ABC = B_i \cup B_j \setminus \{j\}$.

Next $B'_i \cup B'_j \setminus \{j\} \subseteq B_i \cup B_j \setminus \{i\}$ can be shown by assuming for a node $a \in B_i \cup B_j \setminus \{i\}$ it is not in $B'_i \cup B'_j$ and deriving a contradiction. Distinguish the cases $a \in B_i \setminus B_j$, $a \in B_i \cap B_j$ and $a \in B_j \setminus (B_i \setminus \{i\})$.  □

For extended proofs of lemmas in this section see [1]. The proof could also be given in terms of d-separation instead of the axiomatic approach we used. However, for later lemmas this won't be possible so, to keep the style uniform, we did not do it here either. When we perform a swap on two nodes $i$ and $j$ we don't have to adjust any boundary if $i$ and $j$ are not connected. If they are connected only $B_i$ and $B_j$ need to be adjusted and we know $B'_i \subseteq B_i \cup B_j \cup \{j\}$ and $B'_j \subseteq B_i \cup B_j$. Furthermore, the next lemma says if $i \in B_j$ then the nodes of $B_i$ that are not in $B_j$ and $j$ itself will also be in $B'_i$.

**Lemma 3.2** *Let $\theta$, $\theta'$, $i$, $j$, $B_i$, $B_j$, $B'_i$ and $B'_j$ be as defined in Lemma 3.1. Then*

- $B'_a = B_a$   *if $\theta(a) < \theta(i)$*
  *or $\theta(j) < \theta(a)$*
  *or $i \notin B_j \wedge (a = i \vee a = j)$.*
- $B'_i = S \backslash E$   *if $i \in B_j$*
  *in which $S = B_i \cup B_j \cup \{j\}$ and*
  $E = \{a | a \in B_j, \ I(\{i\}, S\backslash\{a\}, U_i\backslash(S\backslash\{a\}))\}$
- $B'_j = S \backslash F$   *if $i \in B_j$*
  *in which $S = B_i \cup B_j\backslash\{i\}$ and*
  $F = \{a | a \in B_j \cup B_i \backslash E,$
  $I(\{j\}, S\backslash\{a\}, U_j\backslash(S\backslash\{a\}))\}$

Sketch of the proof: The case $i \notin B_j$ is trivial. In the case $i \in B_j$ for showing $B'_i$ is a boundary for $i$ one needs to proof $I(i, B'_i, U_i\backslash B'_i)$ does hold and no set $C$ smaller than $B'_i$ exists such that $I(i, C, U_i\backslash C)$.

From Lemma 3.1 derive $B'_i \subseteq B_i \cup B_j \cup \{j\}$. Next show that if a node $a \in B_i \cup \{j\}\backslash B_j$ is not in $B'_i$ a contradiction appears in terms that either $B_i$ or $B_j$ are not boundaries for $i$ and $j$ in $L_\theta$. Then apply intersection several times for the nodes $a \in B_j$ for which $I(i, S\backslash\{a\}, U_i\backslash(S\backslash\{a\}))$ holds to get the final boundary $B'_i = S\backslash E$.

Likewise for $B'_j$ but now regarding Lemma 3.1 for deriving contradictions also. □

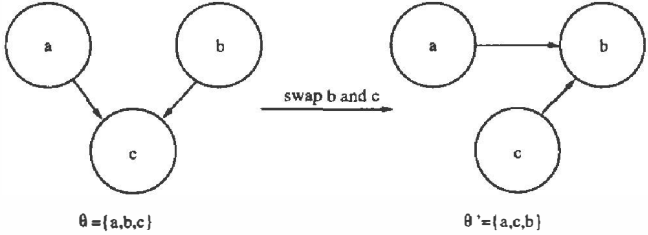

Figure 5: Effect of swap on the model of figure 1

Consider once more the soccer example illustrated by figure 1. So, the set of variables $U = \{a, b, c\}$ with complete independency model $M = \{I(a, \varnothing, b), I(a, \varnothing, c), I(b, \varnothing, c) +$ symmetric statements $\}$ and $\theta = \{a, b, c\}$ results in the left DAG of Figure 5. Swapping $b$ and $c$ results in the DAG on the right. Remark the left DAG represents $I(a, \varnothing, b)$ but the right DAG does not. This example shows the difference with arc reversal.

To apply the strategy described in section 2, we need to know when swapping may be applied in order to obtain a model $M_{\theta'}$ which contains the model $M_\theta$ before the swap.

**Lemma 3.3** *Let $\theta$, $\theta'$, $i$, $j$, $B_i$ and $B_j$ be as defined in Lemma 3.1. Then $M_\theta \subseteq M_{\theta'}$ if and only if $B_i = B_j\backslash\{i\}$ or $i \notin B_j$.*

Sketch of the proof: Assume $a \in B_i$ and $a \notin B_j$ and derive a contradiction by showing that an independency statement that hold in $M_\theta$ cannot hold in $M_{\theta'}$. Likewise for the case $a \notin B_i$ and $a \in B_j$. □

This lemma provides us a simple criterion on the restrictions of our search space. Furthermore, it implies that enlargement of the represented model only is to be expected in cliques of size three or more. In this paper we call a set of nodes a clique only if it contains three or more nodes. To get improvement we have to change at least the ordering in a clique. However, when a clique has incoming arcs from nodes not in the clique restrictions can arise. The term *restriction* will be used for a pair of nodes that cannot be swapped without destroying some independencies i.e. a pair of nodes $i$, $j$ with $i \in B_j$ and $B_i \neq B_j\backslash\{i\}$. A clique $Cl$ is called *restricted by another clique* if a node $a \in Cl$ is restricted by a node $b \notin Cl$ and a clique $Cl'$ exists such that $\{a, b\} \subset Cl'$. In the next sections we will see how we can enlarge the search space by removing restrictions.

## 4 The reversal operator

One way to remove a restriction is by *reversal*: rearranging the ordering such that the restriction on the lowest numbered node of a clique is removed. This can be done when the set of ancestors of the first node in the clique does not contain a head-to-head node.

In Figure 6 we can see why it is necessary to apply reversal when we want to follow a path in $\Theta$ with monotonic increasing size of the corresponding independency model. On the left we see a perfect map and on the right a DAG generated from the ordering $\theta = \{d, c, b, a\}$. To end up with an ordering that results in the left DAG $\theta(d)$ must be last. This cannot be reached when only the nodes in the clique $\{a, b, c\}$ itself are swapped.

An ordering $\sigma$ can be defined on a clique such that for $c_1, c_2 \in Cl$ $\sigma(c_1) < \sigma(c_2)$ iff $\theta(c_1) < \theta(c_2)$. In this sense we can speak of the first node in a clique: the node $c \in Cl$ with $\sigma(c) = 1$.

Let $\theta$ be an ordering and $Cl$ the set of nodes of a clique in $D_\theta$. Let $c$ be the first node in the clique and let $A_c$ be the set of ancestors of $c$ including $c$ and $k = |A_c|$.

The first step in the reversal is reordering $\theta$ such that all nodes in $A_c$ are the first $k$ in the ordering. Let $x \in A_c$ be the node with lowest $\theta(x)$. This node $x$ is shifted towards position 1 in the ordering $\theta$. Next node $y \in A_c$ with one but lowest $\theta(y)$ is shifted towards position 2 in $\theta$, etc. until $\theta(c) = k$. During this shifting the boundary of each node remains the same since no swaps need to be made between nodes that are connected.

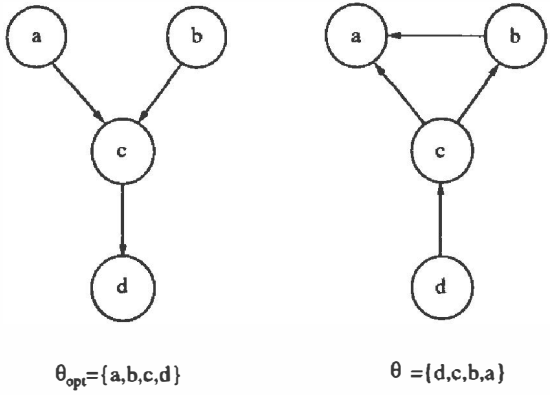

Figure 6: $\theta_{opt}$ will not be reached along a valid path if no reversal takes place.

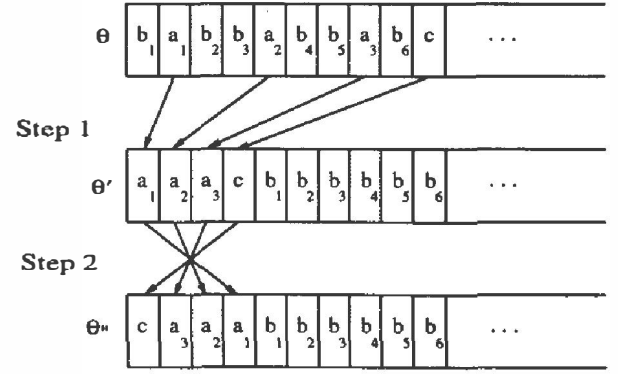

Figure 7: Effect of reversal-operation on an ordering $\theta$ where $a_1$, $a_2$ and $a_3$ are ancestors of $c$ which is the first node in the clique and $b_1 \ldots b_6$ are other nodes.

If $A_c$ contains a tail-to-tail node, i.e. a triple $d \leftarrow e \rightarrow f$ and not $d \rightarrow f$ nor $f \rightarrow d$, then it is necessary to perform the swap on a pair of nodes $e, f$ or $d, f$ in order to remove this tail-to-tail node. This can always be done without destroying independencies if $A_c$ does not contain a head-to-head node.

The second step is to reverse the ordering of the first $k$ nodes. In this case the boundaries are calculated by:

- $B'_a \subseteq \{i | j \in A_c, i \in B_j\}$ if $a \in A_c$ (1)

- $B'_a = B_a$ if $a \in U \backslash A_c$

Here $B_a$ is a boundary in the original ordering and $B'_a$ in the reversed ordering. In Figure 7 the effect of the reversal on the ordering is shown. To get an equality in (1) we have to demand that no clique in $A_c$ can be reordered such that a reduction can be performed.

**Lemma 4.1** *Let $\theta'$ be $\theta$ after reversal and $c$ be the first node in a clique in $D_\theta$ with $A_c$ as set of ancestors including $c$. If $A_c$ does not contain head-to-head nodes or reducible cliques then $M_{\theta'} = M_\theta$.*

Sketch of the proof: Let $D_\theta$ and $D_{\theta'}$ be the corresponding DAGs. $D_{\theta'}$ differs from $D_\theta$ by having all arcs between parents of the first node in the clique turned around. In this way for all nodes $a, b \in U$ if $a$ is connected with $b$ in $D_\theta$ then also in $D_{\theta'}$. Further if $c$ is a head-to-head node in $D_\theta$ with $a$ and $b$ as un-married parents then it is also in $D_{\theta'}$. So $M_\theta = M_{\theta'}$.
□

## 5  The cliquereunion operator

Since we know reduction has to be searched in the reordering of nodes in cliques it is easy to have the nodes of the cliques arranged together in the ordering $\theta$.

An ordering $\sigma$ can be defined on a clique such that for $c_1, c_2 \in Cl$ $\sigma(c_1) < \sigma(c_2)$ iff $\theta(c_1) < \theta(c_2)$. In this sense we can speak of the first node in a clique: the node $c \in Cl$ with $\sigma(c) \leq \sigma(v)$ for all $v \in Cl$.

**Definition 5.1** *Let $\sigma$ be a clique ordering on a clique $Cl$. A free set $F$ is a subset of a clique $Cl$ such that $\forall_{a,b \in F, \sigma(a) = \sigma(b)-1} B_a \subset B_b, \forall_{x \in B_b \backslash B_a \cup \{a\}} B_x = B_b \backslash \{x\}$.*

The next lemma says the nodes in all free sets can be ordered in groups such that no nodes which are not in the free set is in between two nodes of this free set.

**Lemma 5.1** *Let $Cl$ be a clique in a DAG $D_{\theta_0}$. A path $\theta_0, \theta_1, \ldots, \theta_n$ through $\Theta$ exists such that $M_{\theta_0} \subseteq M_{\theta_1} \subseteq \ldots \subseteq M_{\theta_n}$ and every free set $F \subset Cl$ is ordered with $\forall_{a,b \in F} \sigma(a) = \sigma(b) - 1 \Rightarrow \theta_n(a) = \theta_n(b) - 1$.*

Proof: We can arrange this ordering $\theta$ by shifting the nodes in a free set towards the first node in that set. We do this following the ordering $\sigma$. Let $F$ be a free set in $Cl$ and $a, b \in F$ two nodes with $\sigma(a) = \sigma(b) - 1$ and $b$ is shifted towards $a$ using the swap operator. This happens without affecting any reordering of other nodes in $Cl$. In this process we can distinguish three cases:

- $B_b \backslash \{a\} = B_a$: No boundaries are changed since no node $x$ with $\theta(a) < \theta(x) < \theta(b)$ is in $B_b$ thus no swap need to be made with two connected nodes.

- $B_b \backslash \{a\} \neq B_a$ and $x$ with $\theta(x) = \theta(b) - 1$ is not in $B_b$: for the same argument no change of boundaries arise.

- $B_b \backslash \{a\} \neq B_a$ and $x$ with $\theta(x) = \theta(b) - 1$ is in $B_b$: now we perform a swap and

$$B'_x \subseteq B_x \cup \{b\} \wedge B'_b \subseteq B_b \backslash \{x\} \qquad (2)$$

□

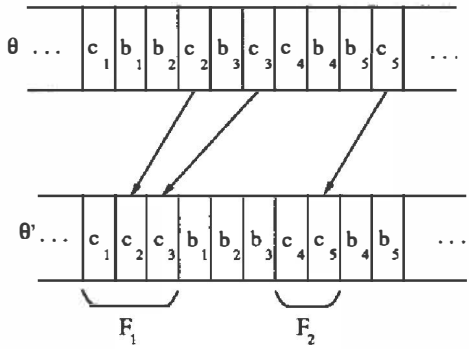

Figure 8: Effect of reunion-operation on an ordering $\theta$ with clique $Cl = F_1 \cup F_2$.

In Figure 8 the effect of cliquereunion on the ordering is shown. The statement for $B'_x$ in (2) can be made an equality if the cliques containing the nodes $b$, $B_x$ and $x$ has already been checked on reducibility.

## 6  The unclique operator

Let a clique $Cl$ be induced by a wrong ordering. Let $X, Z, Y \subset Cl$ be disjoint sets of nodes $X$ and $Y$ not empty and $U_c \subset U$ a subset of nodes with lower ordering than the last node in $Cl$. Assume $I(X, Z \cup U_c, Y)$ holds. Let $R = Cl \backslash X \cup Y \cup Z$. The strategy we apply now is finding an ordering such that at least one arc can be removed in the corresponding DAG. Then we are left with two cliques upon we can apply the same strategy.

Consider the set of variables $U = \{a, b, c, d\}$ with model $M = \{I(b, a, c), I(c, a, b)\}$. Then Figure 9a shows the DAG corresponding to ordering $\theta = \{b, a, d, c\}$. This DAG is a clique actually. The rearranged ordering $\theta' = \{b, a, c, d\}$ has corresponding DAG shown in Figure 9b. The arc between $b$ and $c$ can be removed and the cliques $Cl_1 = \{a, b, d\}$ and $Cl_2 = \{a, c, d\}$ remain.

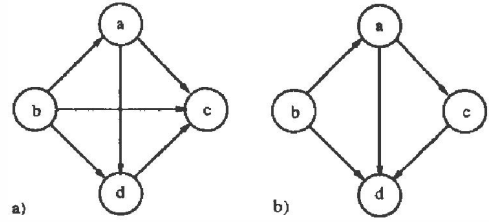

Figure 9: a) Clique  b) Arc $b \rightarrow c$ removed due to ordering

To provide such an ordering the best we can do is just arrange the nodes in the free sets of the clique in any possible ordering if the free sets are not too large. Let $Cl$ be a clique and $F_1, F_2, ..., F_m$ be its free sets. Then there are $\prod_{i=1}^{m}(|F_i|!)$ of such orderings.

## 7  The algorithm

A set $S$ is used which contains all potentially reducible cliques of size three and more. If a clique $Cl \in S$ turns out to be reducible let an arc with tail in $a$ and head in $b$ be an arc which can be removed. Now we have to *split* every clique $Cl' \in S$ containing $a$ and $b$. With splitting is meant that $Cl'$ is removed from $S$ and two new cliques $Cl' \backslash \{a\}$ and $Cl' \backslash \{b\}$ are added to $S$ provided they are of size three or more. After uncliqueing a clique $Cl$ restricted by another clique, it cannot be guaranteed $Cl$ contains no arcs that cannot be removed. After uncliqueing the restricting clique the free sets in $Cl$ may have been enlarged and a new unclique operation on $Cl$ may make some arcs disappear. Therefore, the algorithm uses a set $R$ containing cliques which are restricted by other cliques.

Choosing of cliques are constrained by a special order such that the reversal and reunion operator turn out to be manipulations on orderings and boundaries without the need to use the model $M$. Therefore we define the following ordering on the cliques:

1. Cliques without restriction.
2. Cliques with no cliques in the ancestor-set of the first node which are also in $S$ and not restricted by other cliques in $S$.
3. Cliques not restricted by other cliques in $S$.
4. Remaining cliques.

By uncliqueing the cliques in this order (1) and (2) become equalities. By the way when a clique is uncliqued and two new cliques arise the former ordering must be adapted. To find the optimal ordering we can use Algorithm 1.

The algorithm has complexity exponential with the size of the largest clique due to the unclique operator.

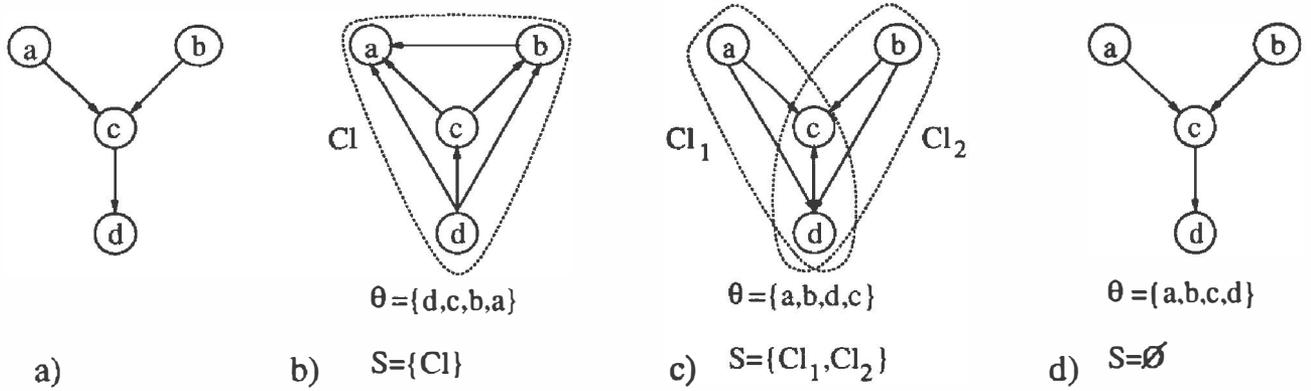

Figure 10: Example of the algorithm application

### Algorithm 1

{S=set of all cliques of size 3 and more}
**input:** Complete independency model $M$,
   Initial ordering $\theta$.
**output:** Optimal ordering $\theta'$.

$R := \emptyset$;
**while** $S \neq \emptyset$ **do**
   $Cl :=$ lowest ordered clique in $S$
   Apply reversal on $Cl$ if possible
   Apply cliquereunion on $Cl$
   Unclique $Cl$
   Split all cliques in $S$ & $R$ containing a removed arc
   $S := S \backslash Cl$
   **if** $Cl$ is restricted by a clique in $S \cup R$ **then**
         $R := R \cup Cl$ **fi**
   **if** $S = \emptyset$ and $R \neq \emptyset$ **then**
      $S := \{Cl | Cl \in R \wedge$ restrictions are removed
            from $Cl\}$ **fi**
**od**

It is commonly assumed BBNs are sparse and do not contain large cliques. By starting with a good ordering, e.g. provided by a maximal cardinality numbering of a chordal graph of this domain [2] or from an expert based on causality, the cliques which arise may be expected to be not too big.

The DAG constructed from the ordering generated by this algorithm is a minimal I-map by method of construction. We conjecture that if $M$ be the complete independency model which is the input of the algorithm and $\theta$ the ordering that is the output then no I-map $D$ of $M$ exists with $M_{D_\theta} \subseteq M_D$ and less arcs than $D_\theta$.

So the DAG resulting from the algorithm has a minimal number of arcs in the sense that no arcs can be removed by rearranging the nodes without having to add some new arcs. However, it is possible a DAG with less arcs exists representing the model. But, this DAG will not represent all independencies represented by the DAG resulting from the algorithm.

We show some examples of the application of the algorithm.

Example 1: Let the DAG in Figure 10a be a perfect map and the DAG in Figure 10b the DAG corresponding to ordering $\theta = \{d, c, b, a\}$. There is only one clique, so we apply the unclique operation on this clique and get the DAG in Figure 10c. The arc between node $a$ and node $b$ is removed thus, the clique $Cl = \{a, b, c, d\}$ of original DAG is split. This splitting results in two cliques $Cl_1 = \{a, c, d\}$ and $Cl_2 = \{b, c, d\}$ which are restricted by each other. It does not matter which clique is uncliqued first. Uncliqueing one of these two cliques will result in the DAG in Figure 10d which is the perfect map of Figure 10a again.

Example 2: Let the complete independency model $M$ contain all statements represented by the DAG in Figure 11a plus $I(f, abc, e)$ plus all statements that can be derived from those statements using the axioms symmetry, weak union, decomposition, contraction and intersection. Then the DAG in Figure 11b is the DAG corresponding to ordering $\theta = \{a, b, c, e, f, d\}$ of this model.

The algorithm starts with set of cliques $S = \{\{a, c, d, e\}, \{a, c, d, f\}, \{b, c, d, f\}\}$. Consider the clique $Cl = \{a, b, d, e\}$. After uncliqueing $Cl$ it turns out that no arc can be removed in $Cl$. In fact $Cl$ consists of four free sets so no swapping is allowed between nodes in $Cl$. $Cl$ is removed from $S$ and, since $Cl$ is restricted by a node in another clique (among others by $f$ in the clique $\{b, c, d, f\}$), $Cl$ is added to $M$. After uncliqueing the other cliques the ordering with corresponding DAG in Figure 11c arises and $S = \emptyset$. The restriction caused by node $f$ on $Cl$ is removed so $Cl$ is added to $S$ again and $Cl$ is uncliqued again, resulting in the DAG in Figure 11a.

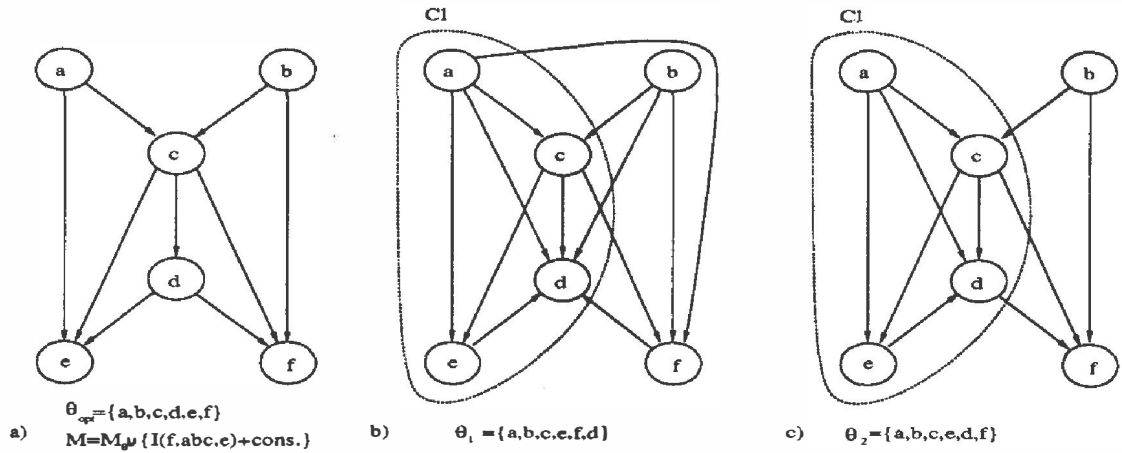

Figure 11: Another example of the algorithm application

## 8 Conclusion

In this paper we have presented an algorithm which generates the structure of a DAG from a data-sample of a probability distribution. The DAG is a minimal I-map of this distribution. Only two assumptions are made namely the distribution is positive definite and any independency statement can be verified i.e. the complete independency model is perfectly known.

The algorithm is based on performing operations on an ordering of the variables which fixes the DAG by the causal input list associated with this ordering and distribution. An operation is performed only if the represented model before the operation is contained in the obtained model. No arcs can be removed by applying the operations on the ordering of resulting DAG unless other arcs have to be added.

We expect that an approach based on manipulating orderings also will be fruitful if the DAG with the minimal number of arcs is searched. Further research has to be done for the case our assumptions are not valid.

## Acknowledgement

This paper has benefited from the helpful suggestions of Linda van der Gaag.